\documentclass[10pt,twocolumn,letterpaper]{article}

\usepackage{cvpr}              
\definecolor{cvprblue}{rgb}{0.21,0.49,0.74}
\usepackage[pagebackref,breaklinks,colorlinks,allcolors=cvprblue]{hyperref}

\def\paperID{5} 
\def\confName{CVPR}
\def\confYear{2026}

\title{SEGAR: Selective Enhancement for Generative Augmented Reality}

\author{ Fanjun Bu \\
Cornell University, Cornell Tech\\
New York, NY, USA \\
{\tt\small fb266@cornell.edu}\\
\and 
Chenyang Yuan \\
Toyota Research Institute \\
Cambridge, MA\\
{\tt\small chenyang.yuan@tri.global}\\
\and 
Hiroshi Yasuda\\
Toyota Research Institute\\
Los Altos, CA\\
{\tt\small hiroshi.yasuda@tri.global}\\
}

\begin{document}
\maketitle
\begin{abstract}
Generative world models offer a compelling foundation for augmented-reality (AR) applications: by predicting future image sequences that incorporate deliberate visual edits, they enable temporally coherent, augmented future frames that can be computed ahead of time and cached, avoiding per-frame rendering from scratch in real time. In this work, we present SEGAR, a preliminary framework that combines a diffusion-based world model with a selective correction stage to support this vision. The world model generates augmented future frames with region-specific edits while preserving others, and the correction stage subsequently aligns safety-critical regions with real-world observations while preserving intended augmentations elsewhere. We demonstrate this pipeline in driving scenarios as a representative setting where semantic region structure is well defined and real-world feedback is readily available. We view this as an early step toward generative world models as practical AR infrastructure, where future frames can be generated, cached, and selectively corrected on demand.
\end{abstract}    
\section{Introduction}
\label{sec:intro}
World models aim to learn predictive representations of the environment that enable agents to imagine future observations, reason about outcomes, and support planning \cite{ding2025understanding, ha2018world}. Recent advances in diffusion-based video generation have substantially improved the fidelity and diversity of predicted future image sequences, leading to strong performance in large-scale generative world models. These models can synthesize temporally and spatially consistent futures from short visual contexts and support controllable generation, making them attractive foundations for autonomous decision-making and interactive perception systems \cite{hu2023gaia, wang2024drivedreamer, gao2024vista, wang2024driving}.

Beyond simulation and planning, we envision a growing role for generative world models in augmented-reality (AR) applications. Unlike conventional AR, which composites virtual assets onto a live camera feed, the Generative AR (GAR) paradigm re-synthesizes the entire view through a generative backbone, producing augmented content as part of the output image itself rather than as a separate overlay \cite{liang2025generative}. This makes generative world models a natural fit: future frames can be generated ahead of time with deliberate augmentations and cached, offering a compelling alternative to per-frame rendering. However, this paradigm places an additional constraint on generative outputs: they must remain spatially and temporally consistent with live reality, since the synthesized view is perceived as reality itself rather than alongside it \cite{guan2023perceptual}.

Despite impressive visual quality, current generative world models may drift from reality in dynamic environments. This drift is compounded in AR, where misalignments can cause perceptual inconsistencies or unsafe interpretations \cite{guan2023perceptual, yamin2024vehicle}. At the same time, AR offers access to real-world observations that can serve as direct corrective feedback. Crucially, because the goal is to generate augmented futures rather than faithful predictions, simply substituting real observations is insufficient; the correction must be selective.

\begin{figure*}[h]
    \centering
    \includegraphics[width=\linewidth]{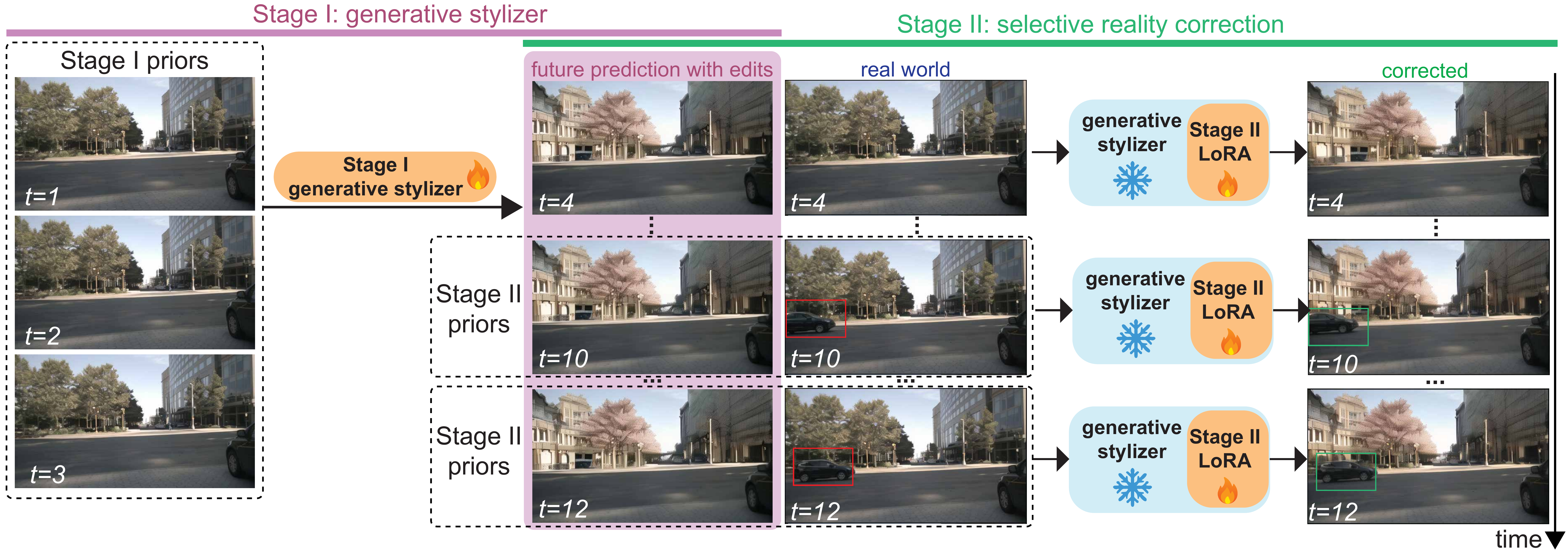}
    \caption{SEGAR system pipeline overview. In Stage I, we train a Vista-based generative stylizer to take three condition frames ($t \in [1,3]$) and output future frames with desired augmented edits ($t \in [4,12]$). In Stage II, the generative stylizer finetuned with LoRA takes the augmented future frame and the corresponding real-world observation as inputs, and outputs a corrected frame in which safety-critical regions are aligned with reality while augmented edits are preserved. In this example, a car that appears in frames 10--12 is absent from the initial prediction because it was not visible in the three condition frames. The corrected output faithfully restores the car while retaining the augmented appearance of non-critical regions.}
    \label{fig:SegarPipeline}
\end{figure*}

In this work, we present SEGAR, a preliminary framework with two stages: a Generative Stylizer that produces augmented future frames with region-specific edits, and a correction stage that aligns safety-critical regions with reality while preserving those augmentations (Fig \ref{fig:SegarPipeline}). This two-stage design offers a practical advantage over real-time video inpainting: stylized future frames can be generated and cached ahead of time, while only the correction stage needs to operate in real time on a per-frame basis, reducing the overall memory and compute footprint at deployment. We demonstrate on driving scenarios as a representative setting and show improved alignment in critical regions without sacrificing augmented edits elsewhere, suggesting a promising direction toward generative world models as practical AR infrastructure.

\section{Related Works}
\subsection{Generative World Models}
Generative world models predict future observations as image sequences, enabling agents to reason about and plan over imagined futures. A representative direction is visual forecasting in driving: Drive-WM generates high-fidelity multiview future videos and demonstrates planning with image-based rewards \cite{wang2024driving}; Vista targets generalizable world modeling with strong fidelity and versatile controllability over long-horizon rollouts \cite{gao2024vista}; and DriveDreamer learns a diffusion-based world model via a two-stage pipeline that captures structured traffic constraints and anticipates future states \cite{wang2024drivedreamer}. Recent work further expands the output space, jointly predicting future RGB and depth \cite{liang2026UniFuture} or synthesizing scene evolution from trajectory prompts \cite{li2025driverse}. Despite this progress, these models operate open-loop and predict what will happen rather than what could happen under deliberate edits, limiting their use in AR settings where generative flexibility and real-world grounding coexist.

\subsection{Generative Augmented Reality}
\citet{liang2025generative} formalize Generative Augmented Reality (GAR) as a paradigm shift from conventional overlay-based AR toward full view re-synthesis driven by a unified generative backbone. Within HCI research, GenAIR articulates a design space for integrating generative AI into AR, focusing on authoring workflows and interaction patterns \cite{hu2023genair}, while \citet{hu2023towards} explores generative image creation in spatial AR through depth-aware diffusion. On the systems side, Osmosis and Dimix demonstrate diffusion-based generation and inpainting in live XR environments \cite{paterakis2025osmosis, taniguchi2023dimix}. At the infrastructure level, \citet{zhang2025generative} proposes edge-side generative models to synthesize XR content under bandwidth and compute constraints.
These works collectively establish the feasibility of generative AR across authoring, rendering, and infrastructure, yet none address the downstream problem of maintaining safety-critical fidelity when generative outputs must persist across frames in a dynamic environment, the core challenge SEGAR is designed to solve.
\section{Methods}
Generating temporally coherent augmented views that remain grounded in live reality calls for a world model that can both anticipate future states and incorporate controllable visual edits. Our approach builds on Vista \cite{gao2024vista}, a diffusion-based driving world model that conditions future frame generation by injecting context frames directly into the denoising process as clean latents. Vista also incorporates dynamics enhancement and structure preservation losses to improve temporal coherence and spatial fidelity in generated sequences. We use Vista as our base world model and extend it in two stages: first, we train it as a \textit{generative stylizer} that generates future frames with region-specific visual edits end-to-end; second, we introduce a correction stage that selectively aligns safety-critical regions of the generated output with real-world observations while preserving the intended edits elsewhere, implemented as a lightweight LoRA finetune over the generative stylizer.

\subsection{Stage I: generative stylizer}
\label{simfuture}
In Stage I, we train Vista to take three consecutive real-world frames as condition images and output future frames with region-specific visual edits. Rather than providing explicit segmentation masks as input at inference time, we train the model end-to-end and rely on the spatial inductive bias of the U-Net backbone to implicitly learn which semantic regions require augmentation and which should remain unchanged \cite{ronneberger2015u, blattmann2023stable}.

\subsubsection{Data Preparation}
\begin{figure}[ht]
    \centering
    \includegraphics[width=\linewidth]{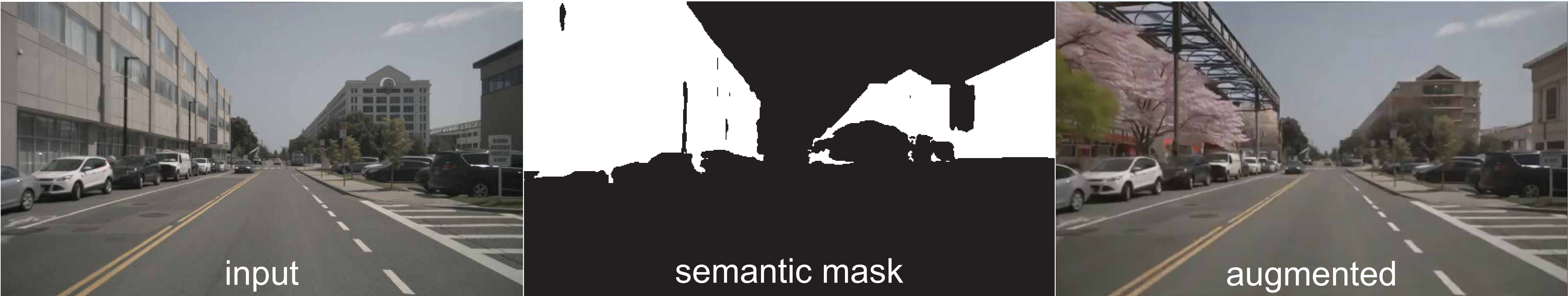}
    \caption{Given an input image sequence, we compute inpainting regions using semantic segmentation. The resulting masks guide VACE's inpainting process to augment static scene elements into a Tokyo-style appearance.}
    \label{fig:StylizePipeline}
\end{figure}

\begin{figure}[h]
    \centering
    \includegraphics[width=\linewidth]{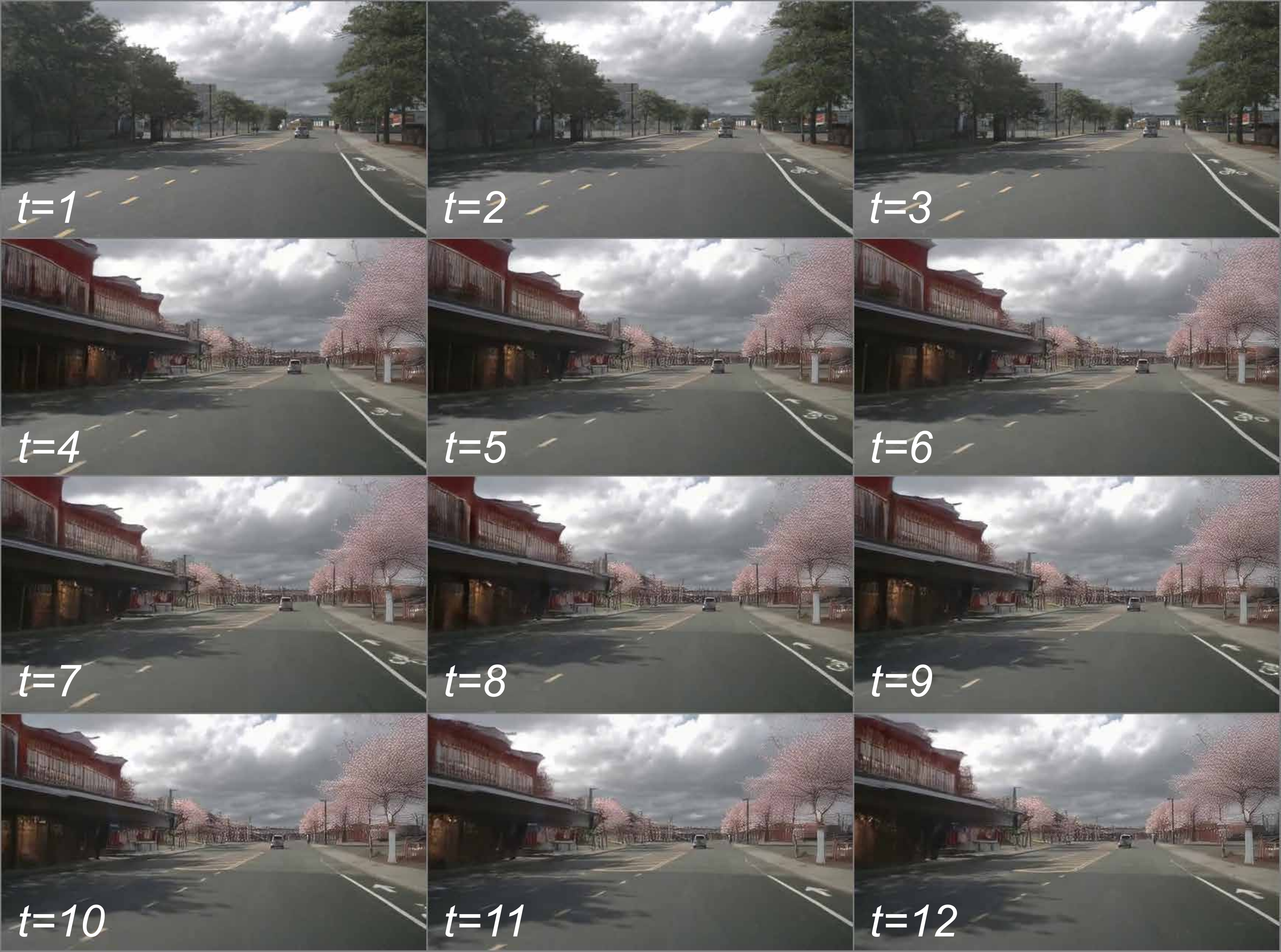}
    \caption{Example training target for Stage I training. The three frames in the top row are condition frames, whose clean latents are injected into the denoising process via dynamic prior injection \cite{gao2024vista}. The remaining nine future frames are generated using VACE with desired visual augmentations.}
    \label{fig:datasetDemo}
\end{figure}

Our training samples are generated based on the nuScenes dataset \cite{caesar2020nuscenes}. To generate training targets, we leverage the VACE framework built on Wan2.1-1.3B \cite{vace, wan2025} to inpaint selected scene regions into a different visual style. In our experiments, we prompt VACE with Tokyo street aesthetics (see Appendix for detailed prompts). To determine which regions to edit, we apply SegFormer to identify non-critical static elements for driving, including building, wall, vegetation, etc. (Fig. \ref{fig:StylizePipeline}) \cite{xie2021segformer}. The resulting masks guide VACE's inpainting process, ensuring that dynamic and safety-critical elements such as road, car, person, and traffic signs and lights remain unchanged (Fig. ~\ref{fig:datasetDemo}). Due to computational constraints, we limit generation to 12 frames, including 3 conditioning frames.

\subsubsection{Stage I Training}
We follow Vista's first training phase with minor modifications \cite{gao2024vista}. Unlike the original Vista paper, the first three frames always serve as condition inputs via dynamic prior injection, and the targets are the corresponding nine VACE-augmented future frames. We reuse the dynamics enhancement and structure preservation losses in addition to the standard diffusion loss, following the same loss formulation. No additional input signals, such as trajectories or steering, are used. Like Vista, we also initialize from pretrained Stable Video Diffusion (SVD) and train until coherent augmentation and motion properties emerge in sampled outputs. All images are processed at a resolution of $320 \times 576$ due to computational constraints. Our training recipe closely follows Vista's phase one setup, leveraging DeepSpeed Stage 2 for distributed training. 

\subsection{Stage II: Selective Reality Correction}
Given sampled future frames from Stage I, Stage II corrects safety-critical regions to align with real-world observations while preserving intended augmentations elsewhere. Unlike Stage I, which processes all 12 frames jointly to leverage temporal context, Stage II operates on each frame independently, as the correction objective is defined per-frame based on the corresponding real-world observation.

\subsubsection{Input Formulation}
Each training example in Stage II is a triplet $(x,\, x^{\text{aug}},\, \hat{x})$, where $x$ is the real-world observation, $x^{\text{aug}}$ is the corresponding augmented future frame sampled from the model trained in Stage I, and $\hat{x}$ is the model output to be learned, which is initialized by copy and concatenate $x^{\text{aug}}$. In practice, although Stage I generates nine future frames, we only use the last frame to form triplets as a practical simplification. The latents of $x$ and $x^{\text{aug}}$ are injected as clean latents by replacing their corresponding noisy latents in the denoising process, mirroring Vista's dynamic prior injection mechanism~\cite{gao2024vista}.

\subsubsection{Conditioning Design}
Stage II builds on SVD's conditioning architecture, which provides two complementary pathways: 
(1) a VAE-latent spatial path, where the conditioning frame is encoded and concatenated at the UNet input to provide pixel-aligned spatial grounding, and
(2) a CLIP embedding path, injected via cross-attention, supplying global semantic context that guides scene identity, structure, and motion coherence.
In the original Vista formulation, both paths receive the same conditioning frame. For Stage II, we decouple them: the VAE path receives the real observation $x$, providing a strong spatial anchor that directly enforces reality grounding in critical regions, while the CLIP path receives the augmented frame $x^{\text{aug}}$, aligning the global semantic anchor with the target distribution of the augmented output.

\subsubsection{Loss Function}
We define a spatially masked reconstruction loss in the latent space that enforces different objectives across image regions:
\begin{equation}
\begin{split}
    \mathcal{L} = \frac{1}{N} \sum_{i} w(\sigma) \cdot \bigl[
    & m_{\text{crit}}^{(i)} \left(\hat{x}^{(i)}_{latent} - x^{(i)}_{latent}\right)^2 \\
    + \; & m_{\text{aug}}^{(i)} \left(\hat{x}^{(i)}_{latent} - x^{\text{aug}(i)}_{latent}\right)^2
    \bigr]
\end{split}
\end{equation}
where $i$ indexes over spatial locations and latent channels, $N$ is the total number of elements, $w(\sigma)$ is the diffusion loss weighting inherited from Vista, $m_{\text{crit}}$ is the safety-critical mask enforcing fidelity to the real observation $x$, and $m_{\text{aug}}$ is the augmentation mask enforcing preservation of the stylistic edits in $x^{\text{aug}}$.

\begin{figure}[h]
    \centering
    \includegraphics[width=\linewidth]{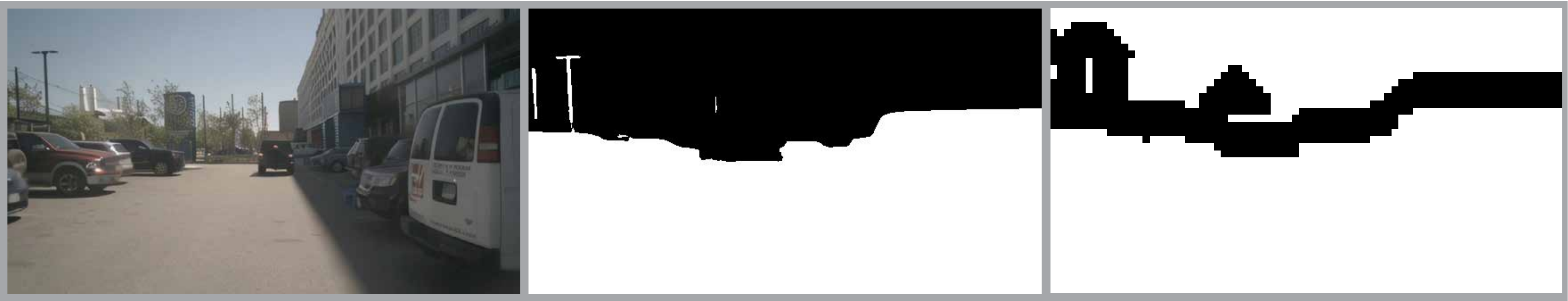}
    \caption{(Left) Real-world image. (Middle) Semantic segmentation of safety-critical regions. (Right) The Buffer zone is introduced by dilating the safety-critical mask, shown in black. Loss is not computed in the buffer zone.}
    \label{fig:lossMask}
\end{figure}

\begin{figure}[h]
    \centering
    \includegraphics[width=\linewidth]{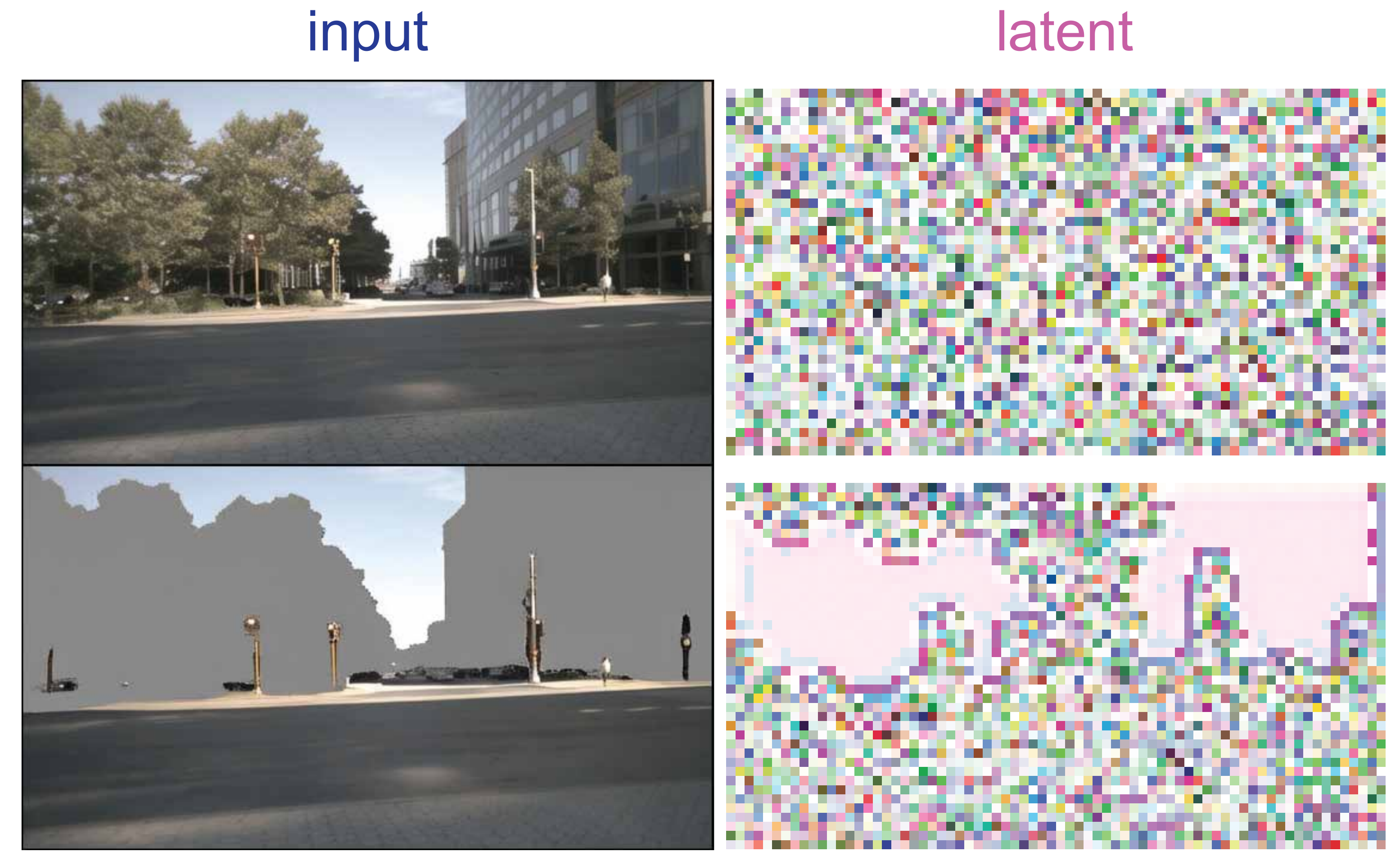}
    \caption{Visualization of latent representations for the real observation before (top) and after (bottom) masking (masked regions inpainted in gray), showing that spatial structure is preserved in the latent space (right) and validating the use of mask downsampling for region-specific loss computation.}
    \label{fig:latentVisual}
\end{figure}

The safety-critical mask $m_{\text{crit}}$ is precomputed offline using SegFormer applied to the real observation $x$, selecting semantic categories corresponding to roads, vehicles, pedestrians, and traffic signs (Fig.~\ref{fig:lossMask}). The augmentation mask $m_{\text{aug}}$ is initialized as the complement of $m_{\text{crit}}$, identifying static non-critical elements such as buildings and vegetation. A buffer zone is then introduced by morphologically dilating $m_{\text{crit}}$ by 40 pixels at the original image resolution using a structuring element that extends upward and laterally but not downward, and subtracting the resulting ring from $m_{\text{aug}}$. This asymmetric dilation is geometrically motivated by driving scene structure: the boundary between safety-critical regions and augmentable regions is predominantly horizontal, and preventing downward expansion avoids consuming road supervision area while pushing the buffer into the augmentation region above. Both masks are subsequently downsampled to the latent spatial dimensions using nearest-neighbour interpolation to preserve binary values. Pixels within the buffer ring carry no reconstruction loss, allowing the model to rely on its pretrained prior to harmonize the transition between regions (Fig.~\ref{fig:latentVisual}, Fig.~\ref{fig:lossMask}).

\subsubsection{Stage II Training}
We finetune the Stage I generative stylizer using rank-16 LoRA adapters~\cite{hu2022lora}. Adapters are applied exclusively to the spatial attention layers of the U-Net. Critically, we do not apply LoRA to the temporal attention layers. Semantic region information is introduced implicitly through the spatially masked loss rather than through explicit conditioning signals, allowing correction to be applied at inference without requiring segmentation masks as input. Offset noise is disabled during training as it would introduce a training-inference mismatch for the single-frame correction objective. All other hyperparameters follow Vista's original configuration for phase two.

\section{Result}
\begin{figure}[ht]
    \centering
    \includegraphics[width=\linewidth]{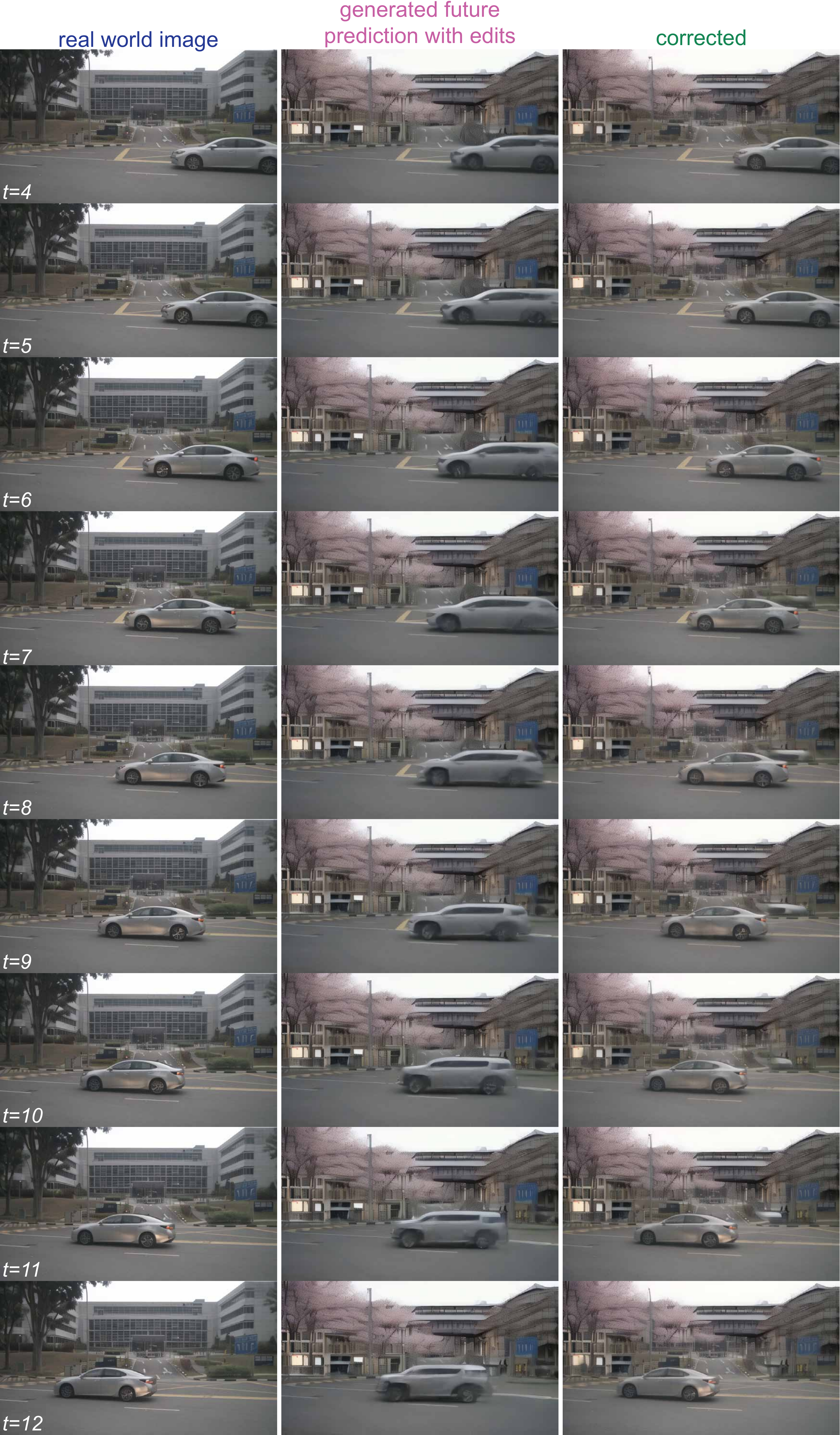}
    \caption{Side-by-side comparison for a sequence of future predictions and their corresponding corrected images.}
    \label{fig:video_result}
\end{figure}

\begin{figure*}[ht]
    \centering
    \includegraphics[width=\linewidth]{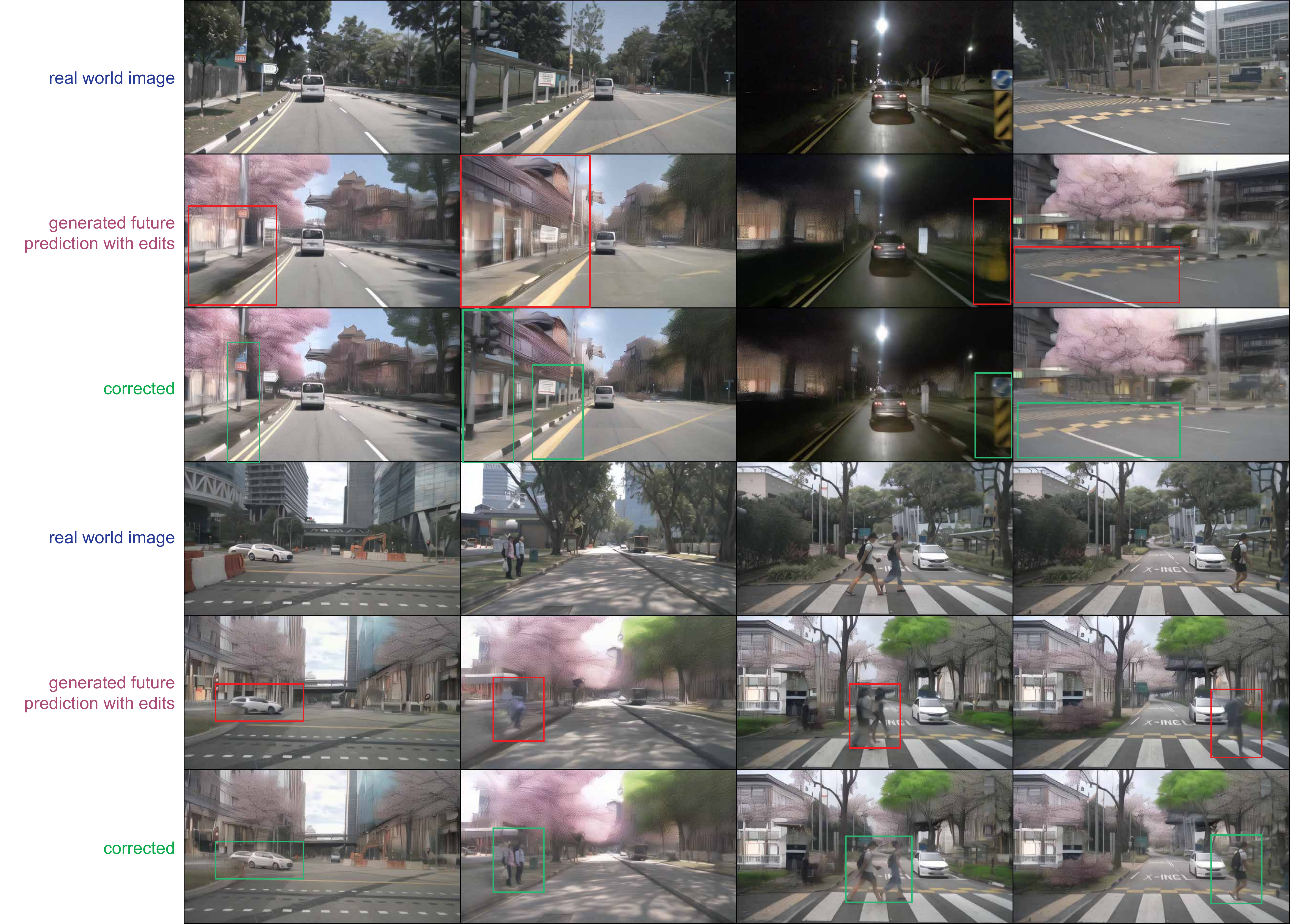}
    \caption{Side-by-side comparison shows that our model successfully corrects the semantically important region while preserving edits in various scenarios (pedestrians, road signs, other vehicles, etc.).}
    \label{fig:result}
\end{figure*}

Qualitatively, as shown in Fig. ~\ref{fig:video_result} and Fig. \ref{fig:result}, safety-critical regions are corrected to align with reality while the augmented edits are preserved. Misplaced pedestrians, vehicles, and road signs are faithfully restored to match real-world observations, while buildings and trees retain the intended stylistic appearance.

\begin{table}[h]
\centering
\renewcommand{\arraystretch}{1.2}
\setlength{\tabcolsep}{5pt}
\begin{tabular}{lcccc}
\toprule
 & \multicolumn{2}{c}{\textbf{Critical}} & \multicolumn{1}{c}{\textbf{Augmented}} \\
\cmidrule(lr){2-3} \cmidrule(lr){4-4}
 & Real vs Corr. & Real vs Aug. & Aug. vs Corr. \\
\midrule
\textbf{SSIM} $\uparrow$ & \textbf{0.9431} & 0.7698 & 0.8662 \\
\textbf{LPIPS} $\downarrow$ & 0.2854 & 0.3968 & \textbf{0.1295} \\
\bottomrule
\end{tabular}
\caption{Similarity metrics across safety-critical (road) and augmentation (building) regions. Real vs.\ Aug.\ serves as the baseline drift introduced by Stage I. Stage II substantially improves alignment with reality in safety-critical regions (SSIM $0.770 \to 0.943$, LPIPS $0.397 \to 0.285$) while preserving the intended augmentation elsewhere (SSIM $0.866$, LPIPS $0.130$).}
\label{tab:region_metrics}
\end{table}

We evaluate our approach using a slightly modified SSIM and LPIPS computed separately over two semantically defined regions. For each test example, we form a triplet consisting of the real-world observation $x$, the augmented future frame $x^{\text{aug}}$, and the corrected output $\hat{x}$. A semantic segmentation mask is computed on $x$ using SegFormer to identify safety-critical regions, with the complement serving as the 
augmentation region.

SSIM is computed using a masked Gaussian-windowed formulation. Rather than applying a standard normalized Gaussian window, local statistics are computed only over 
masked pixels:
\begin{equation}
    \mu_x = \frac{\sum_{k,l} w_{kl} \cdot m_{i+k,j+l} \cdot x_{i+k,j+l}}
                 {\sum_{k,l} w_{kl} \cdot m_{i+k,j+l}}
\end{equation}
where $w$ is an $11 \times 11$ Gaussian kernel with $\sigma = 1.5$ and $m_{ij} \in \{0,1\}$ is the binary mask. Variance and covariance terms are computed analogously. A pixel at location $(i,j)$ is included in the final mean only if the Gaussian-weighted fraction of its window that falls inside the mask meets a minimum threshold $\tau = 0.8$:
\begin{equation}
    \frac{\sum_{k,l} w_{kl} \cdot m_{i+k,\,j+l}}
         {\sum_{k,l} w_{kl}} \geq \tau
\end{equation}
LPIPS is computed by cropping images to the bounding box of the relevant mask region before evaluation.

In the safety-critical region, we measure similarity between $x$ and $\hat{x}$, as the corrected output should closely match reality there. In the augmentation region, we measure similarity between $x^{\text{aug}}$ and $\hat{x}$, as the corrected output should preserve the intended stylistic edits. We also report similarity between $x$ and $x^{\text{aug}}$ in the safety-critical region as a baseline, reflecting the drift introduced in Stage I.

\begin{figure*}
    \centering
    \includegraphics[width=\linewidth]{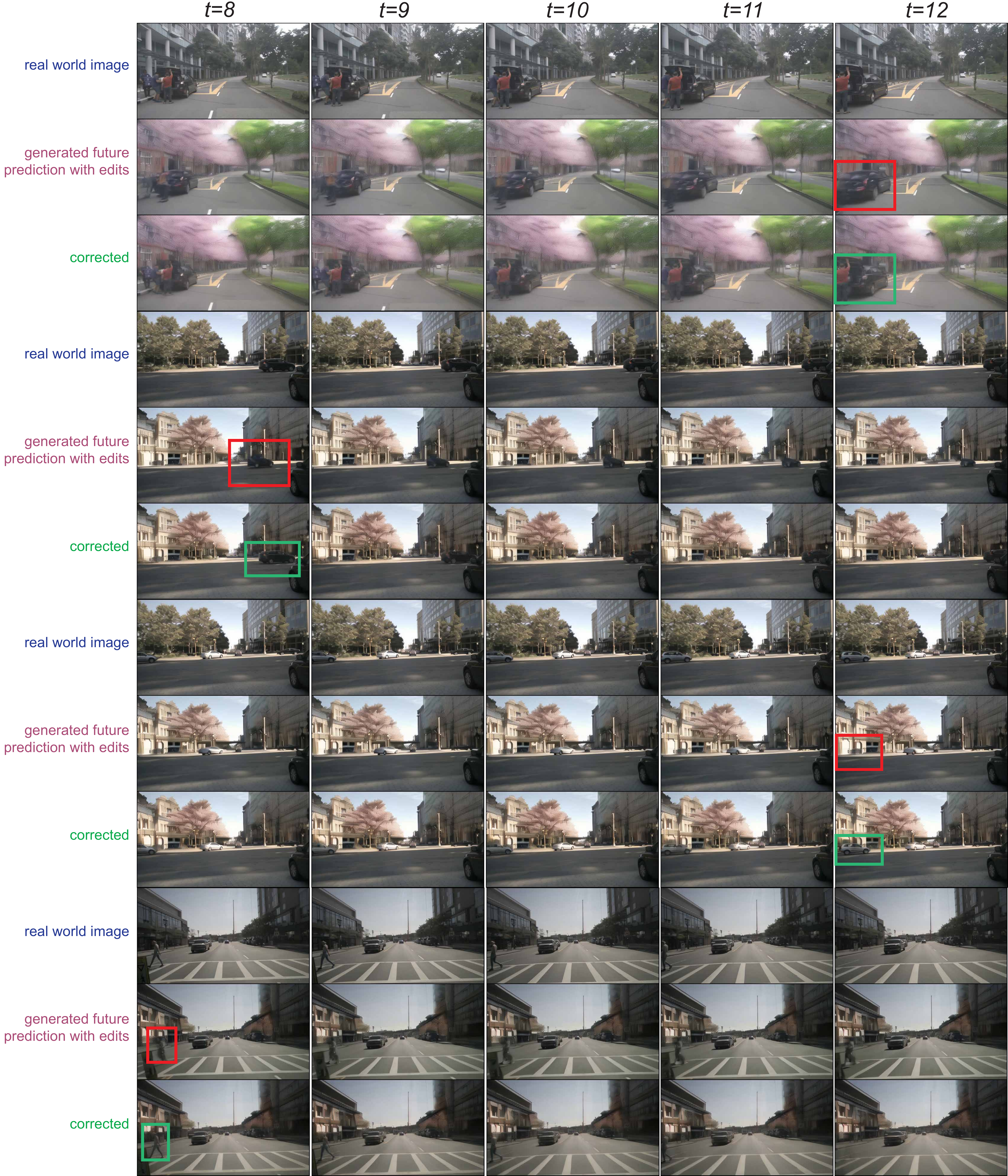}
    \caption{More demos showing how the correction stage corrects safety-critical regions of the image, such as cars or pedestrians.}
    \label{fig:placeholder}
\end{figure*}

Table~\ref{tab:region_metrics} reports results across both regions. In the safety-critical region, the corrected output achieves an SSIM of 0.943 against the real observation, compared to 0.770 for the Stage I augmented frame, indicating that Stage II successfully restores fidelity in regions that must align with reality. Notably, since our loss operates at the pixel level in latent space without explicit structural constraints, the high SSIM suggests that the pretrained diffusion prior plays an important role in maintaining structural coherence during correction. LPIPS similarly decreases from 0.397 to 0.285, though the relatively higher LPIPS compared to SSIM suggests that some perceptual differences in texture and appearance remain beyond what pixel-level supervision can resolve \cite{zhang2018unreasonable}. In the augmentation region, the corrected output maintains an SSIM of 0.866 and an LPIPS of 0.130 against the Stage I augmented frame, suggesting that the intended visual edits are largely preserved. Taken together, the results show that Stage II improves reality alignment in safety-critical regions without substantially degrading the augmentation elsewhere.

\subsection{Potential Applications}
Although we demonstrate SEGAR in driving scenarios, the framework is motivated by and broadly applicable to generative AR settings. In GAR systems, where the entire view is re-synthesized by a generative backbone rather than composited from discrete assets \cite{liang2025generative}, future frames could be generated with deliberate edits, cached ahead of time, and selectively corrected against incoming real-world observations as the user moves through the environment. Beyond AR, the selective correction mechanism applies to any setting where a generative model must remain grounded in real-world feedback while preserving intentional modifications, such as robotic perception or simulation-to-real transfer.

\subsection{Limitations}
\begin{figure}[h]
    \centering
    \includegraphics[width=\linewidth]{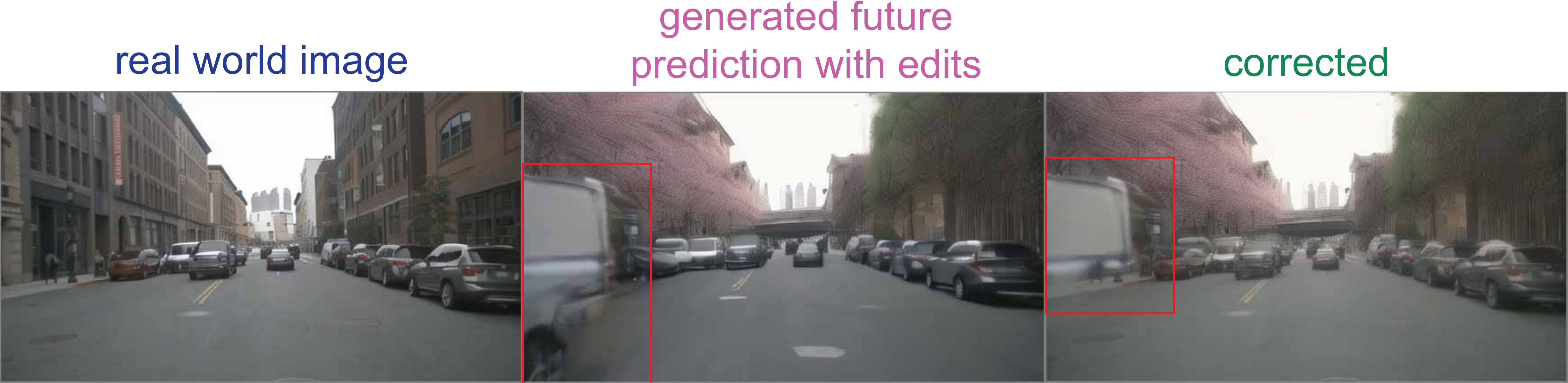}
    \caption{Failure case illustrating object-level inconsistency at region boundaries.}
    \label{fig:limitation}
\end{figure}
Our current work remains a proof-of-concept, and we observe several notable limitations that should be addressed in future work.

Currently, the model lacks explicit object-level understanding and operates at the region level defined by the real-world observation. When objects span across the correction and augmentation regions, conflicting reconstruction objectives are applied to the same object, producing inconsistent corrections. An example is shown in Fig.~\ref{fig:limitation}, where a truck is only partially corrected, leaving it fragmented across the two regions in the output. 

A related limitation concerns the buffer zone between the two mask regions. While temporal consistency is largely preserved elsewhere, since Stage I outputs and real-world observations are both temporally coherent, the buffer zone receives no reconstruction supervision, and no ground truth target exists for these transition regions. Since Stage II operates on a per-frame basis without any explicit temporal constraint, the boundary blending may vary across frames, producing flickering artifacts at region boundaries. This is a limitation inherent to the current design rather than a failure of training.


The current framework also cannot be applied autoregressively. Vista supports long-horizon prediction by using the last few generated frames as dynamic priors for the next rollout step. However, Stage I is trained to condition on real-world frames, and the corrected outputs from Stage II are a hybrid of real and augmented content that falls outside this training distribution. Feeding corrected frames back as condition inputs would therefore introduce a distribution mismatch, limiting the framework to single-step prediction rather than continuous, long-horizon rollouts. A promising direction for future work would be to use the corrected output from Stage II as an additional condition image, allowing the style established in the previous step to propagate forward into subsequent predictions.

Finally, the generative stylizer is trained on a dataset augmented into a single target style, meaning that producing a different visual style requires regenerating the entire training dataset with a new VACE prompt and retraining the model. Future work could address this by incorporating explicit style conditioning signals such as text prompts or reference images, enabling the generative stylizer to apply diverse edits at inference time without retraining.

\section{Conclusion}

We presented SEGAR, a preliminary two-stage framework that combines a diffusion-based world model for generating edited future frames with a selective correction stage that aligns safety-critical regions with real-world observations while preserving intended stylistic edits elsewhere. Experiments on driving scenarios demonstrate improved alignment in critical regions without sacrificing generative flexibility, suggesting that the proposed pipeline is a promising step toward generative world models as practical AR infrastructure, where future frames can be generated with deliberate edits, cached, and selectively corrected against real-world observations on demand.

{
    \small
    \bibliographystyle{ieeenat_fullname}
    \bibliography{main}

\begin{thebibliography}{24}
\providecommand{\natexlab}[1]{#1}
\providecommand{\url}[1]{\texttt{#1}}
\expandafter\ifx\csname urlstyle\endcsname\relax
  \providecommand{\doi}[1]{doi: #1}\else
  \providecommand{\doi}{doi: \begingroup \urlstyle{rm}\Url}\fi

\bibitem[Blattmann et~al.(2023)Blattmann, Dockhorn, Kulal, Mendelevitch, Kilian, Lorenz, Levi, English, Voleti, Letts, et~al.]{blattmann2023stable}
Andreas Blattmann, Tim Dockhorn, Sumith Kulal, Daniel Mendelevitch, Maciej Kilian, Dominik Lorenz, Yam Levi, Zion English, Vikram Voleti, Adam Letts, et~al.
\newblock Stable video diffusion: Scaling latent video diffusion models to large datasets.
\newblock \emph{arXiv preprint arXiv:2311.15127}, 2023.

\bibitem[Caesar et~al.(2020)Caesar, Bankiti, Lang, Vora, Liong, Xu, Krishnan, Pan, Baldan, and Beijbom]{caesar2020nuscenes}
Holger Caesar, Varun Bankiti, Alex~H Lang, Sourabh Vora, Venice~Erin Liong, Qiang Xu, Anush Krishnan, Yu Pan, Giancarlo Baldan, and Oscar Beijbom.
\newblock nuscenes: A multimodal dataset for autonomous driving.
\newblock In \emph{Proceedings of the IEEE/CVF conference on computer vision and pattern recognition}, pages 11621--11631, 2020.

\bibitem[Ding et~al.(2025)Ding, Zhang, Shang, Zhang, Zong, Feng, Yuan, Su, Li, Sukiennik, et~al.]{ding2025understanding}
Jingtao Ding, Yunke Zhang, Yu Shang, Yuheng Zhang, Zefang Zong, Jie Feng, Yuan Yuan, Hongyuan Su, Nian Li, Nicholas Sukiennik, et~al.
\newblock Understanding world or predicting future? a comprehensive survey of world models.
\newblock \emph{ACM Computing Surveys}, 58\penalty0 (3):\penalty0 1--38, 2025.

\bibitem[Gao et~al.(2024)Gao, Yang, Chen, Chitta, Qiu, Geiger, Zhang, and Li]{gao2024vista}
Shenyuan Gao, Jiazhi Yang, Li Chen, Kashyap Chitta, Yihang Qiu, Andreas Geiger, Jun Zhang, and Hongyang Li.
\newblock Vista: A generalizable driving world model with high fidelity and versatile controllability.
\newblock In \emph{Advances in Neural Information Processing Systems (NeurIPS)}, 2024.

\bibitem[Guan et~al.(2023)Guan, Penner, Hegland, Letham, and Lanman]{guan2023perceptual}
Phillip Guan, Eric Penner, Joel Hegland, Benjamin Letham, and Douglas Lanman.
\newblock Perceptual requirements for world-locked rendering in ar and vr.
\newblock In \emph{SIGGRAPH Asia 2023 Conference Papers}, pages 1--10, 2023.

\bibitem[Ha and Schmidhuber(2018)]{ha2018world}
David Ha and J{\"u}rgen Schmidhuber.
\newblock World models.
\newblock \emph{arXiv preprint arXiv:1803.10122}, 2\penalty0 (3):\penalty0 440, 2018.

\bibitem[Hu et~al.(2023{\natexlab{a}})Hu, Russell, Yeo, Murez, Fedoseev, Kendall, Shotton, and Corrado]{hu2023gaia}
Anthony Hu, Lloyd Russell, Hudson Yeo, Zak Murez, George Fedoseev, Alex Kendall, Jamie Shotton, and Gianluca Corrado.
\newblock Gaia-1: A generative world model for autonomous driving.
\newblock \emph{arXiv preprint arXiv:2309.17080}, 2023{\natexlab{a}}.

\bibitem[Hu et~al.(2022)Hu, Shen, Wallis, Allen-Zhu, Li, Wang, Wang, Chen, et~al.]{hu2022lora}
Edward~J Hu, Yelong Shen, Phillip Wallis, Zeyuan Allen-Zhu, Yuanzhi Li, Shean Wang, Liang Wang, Weizhu Chen, et~al.
\newblock Lora: Low-rank adaptation of large language models.
\newblock \emph{Iclr}, 1\penalty0 (2):\penalty0 3, 2022.

\bibitem[Hu et~al.(2023{\natexlab{b}})Hu, Hu, and Quigley]{hu2023towards}
Yongquan Hu, Wen Hu, and Aaron Quigley.
\newblock Towards using generative ai for facilitating image creation in spatial augmented reality.
\newblock In \emph{2023 IEEE International Symposium on Mixed and Augmented Reality Adjunct (ISMAR-Adjunct)}, pages 441--443. IEEE, 2023{\natexlab{b}}.

\bibitem[Hu et~al.(2023{\natexlab{c}})Hu, Zhang, and Quigley]{hu2023genair}
Yongquan Hu, Dawen Zhang, and Aaron Quigley.
\newblock Genair: exploring design factor of employing generative ai for augmented reality.
\newblock In \emph{Proceedings of the 2023 ACM Symposium on Spatial User Interaction}, pages 1--3, 2023{\natexlab{c}}.

\bibitem[Jiang et~al.(2025)Jiang, Han, Mao, Zhang, Pan, and Liu]{vace}
Zeyinzi Jiang, Zhen Han, Chaojie Mao, Jingfeng Zhang, Yulin Pan, and Yu Liu.
\newblock Vace: All-in-one video creation and editing.
\newblock In \emph{Proceedings of the IEEE/CVF International Conference on Computer Vision}, pages 17191--17202, 2025.

\bibitem[Li et~al.(2025)Li, Wu, Yang, Xu, Zhang, Liang, Wan, and Wang]{li2025driverse}
Xiaofan Li, Chenming Wu, Zhao Yang, Zhihao Xu, Yumeng Zhang, Dingkang Liang, Ji Wan, and Jun Wang.
\newblock Driverse: Navigation world model for driving simulation via multimodal trajectory prompting and motion alignment.
\newblock In \emph{Proceedings of the 33rd ACM International Conference on Multimedia}, pages 9753--9762, 2025.

\bibitem[Liang et~al.(2025)Liang, Zheng, Zeng, Tan, Lyu, Zheng, Li, Weng, Shi, and Zhang]{liang2025generative}
Chen Liang, Jiawen Zheng, Yufeng Zeng, Yi Tan, Hengye Lyu, Yuhui Zheng, Zisu Li, Yueting Weng, Jiaxin Shi, and Hanwang Zhang.
\newblock Generative augmented reality: Paradigms, technologies, and future applications.
\newblock \emph{arXiv preprint arXiv:2511.16783}, 2025.

\bibitem[Liang et~al.(2026)Liang, Zhang, Zhou, Tu, Feng, Li, Zhang, Du, Tan, and Bai]{liang2026UniFuture}
Dingkang Liang, Dingyuan Zhang, Xin Zhou, Sifan Tu, Tianrui Feng, Xiaofan Li, Yumeng Zhang, Mingyang Du, Xiao Tan, and Xiang Bai.
\newblock Unifuture: A 4d driving world model for future generation and perception.
\newblock In \emph{IEEE International Conference on Robotics and Automation}, 2026.

\bibitem[Paterakis and Manoudaki(2025)]{paterakis2025osmosis}
Iason Paterakis and Nefeli Manoudaki.
\newblock Osmosis: Generative ai and xr for the real-time transformation of urban architectural environments.
\newblock \emph{International Journal of Architectural Computing}, 23\penalty0 (4):\penalty0 821--836, 2025.

\bibitem[Ronneberger et~al.(2015)Ronneberger, Fischer, and Brox]{ronneberger2015u}
Olaf Ronneberger, Philipp Fischer, and Thomas Brox.
\newblock U-net: Convolutional networks for biomedical image segmentation.
\newblock In \emph{International Conference on Medical image computing and computer-assisted intervention}, pages 234--241. Springer, 2015.

\bibitem[Taniguchi(2023)]{taniguchi2023dimix}
Daiki Taniguchi.
\newblock Dimix: A cross-dimensional mixed reality system based on latent diffusion model.
\newblock In \emph{ACM SIGGRAPH 2023 Immersive Pavilion}, pages 1--2. 2023.

\bibitem[Wan et~al.(2025)Wan, Wang, Ai, Wen, Mao, Xie, Chen, Yu, Zhao, Yang, Zeng, Wang, Zhang, Zhou, Wang, Chen, Zhu, Zhao, Yan, Huang, Feng, Zhang, Li, Wu, Chu, Feng, Zhang, Sun, Fang, Wang, Gui, Weng, Shen, Lin, Wang, Wang, Zhou, Wang, Shen, Yu, Shi, Huang, Xu, Kou, Lv, Li, Liu, Wang, Zhang, Huang, Li, Wu, Liu, Pan, Zheng, Hong, Shi, Feng, Jiang, Han, Wu, and Liu]{wan2025}
Team Wan, Ang Wang, Baole Ai, Bin Wen, Chaojie Mao, Chen-Wei Xie, Di Chen, Feiwu Yu, Haiming Zhao, Jianxiao Yang, Jianyuan Zeng, Jiayu Wang, Jingfeng Zhang, Jingren Zhou, Jinkai Wang, Jixuan Chen, Kai Zhu, Kang Zhao, Keyu Yan, Lianghua Huang, Mengyang Feng, Ningyi Zhang, Pandeng Li, Pingyu Wu, Ruihang Chu, Ruili Feng, Shiwei Zhang, Siyang Sun, Tao Fang, Tianxing Wang, Tianyi Gui, Tingyu Weng, Tong Shen, Wei Lin, Wei Wang, Wei Wang, Wenmeng Zhou, Wente Wang, Wenting Shen, Wenyuan Yu, Xianzhong Shi, Xiaoming Huang, Xin Xu, Yan Kou, Yangyu Lv, Yifei Li, Yijing Liu, Yiming Wang, Yingya Zhang, Yitong Huang, Yong Li, You Wu, Yu Liu, Yulin Pan, Yun Zheng, Yuntao Hong, Yupeng Shi, Yutong Feng, Zeyinzi Jiang, Zhen Han, Zhi-Fan Wu, and Ziyu Liu.
\newblock Wan: Open and advanced large-scale video generative models.
\newblock \emph{arXiv preprint arXiv:2503.20314}, 2025.

\bibitem[Wang et~al.(2024{\natexlab{a}})Wang, Zhu, Huang, Chen, Zhu, and Lu]{wang2024drivedreamer}
Xiaofeng Wang, Zheng Zhu, Guan Huang, Xinze Chen, Jiagang Zhu, and Jiwen Lu.
\newblock Drivedreamer: Towards real-world-drive world models for autonomous driving.
\newblock In \emph{European conference on computer vision}, pages 55--72. Springer, 2024{\natexlab{a}}.

\bibitem[Wang et~al.(2024{\natexlab{b}})Wang, He, Fan, Li, Chen, and Zhang]{wang2024driving}
Yuqi Wang, Jiawei He, Lue Fan, Hongxin Li, Yuntao Chen, and Zhaoxiang Zhang.
\newblock Driving into the future: Multiview visual forecasting and planning with world model for autonomous driving.
\newblock In \emph{Proceedings of the IEEE/CVF Conference on Computer Vision and Pattern Recognition}, pages 14749--14759, 2024{\natexlab{b}}.

\bibitem[Xie et~al.(2021)Xie, Wang, Yu, Anandkumar, Alvarez, and Luo]{xie2021segformer}
Enze Xie, Wenhai Wang, Zhiding Yu, Anima Anandkumar, Jose~M Alvarez, and Ping Luo.
\newblock Segformer: Simple and efficient design for semantic segmentation with transformers.
\newblock In \emph{Neural Information Processing Systems (NeurIPS)}, 2021.

\bibitem[Yamin et~al.(2024)Yamin, Park, and Kim]{yamin2024vehicle}
Putra~AR Yamin, Jaehyun Park, and Hyun~K Kim.
\newblock In-vehicle human--machine interface guidelines for augmented reality head-up displays: A review, guideline formulation, and future research directions.
\newblock \emph{Transportation Research Part F: Traffic Psychology and Behaviour}, 104:\penalty0 266--285, 2024.

\bibitem[Zhang et~al.(2018)Zhang, Isola, Efros, Shechtman, and Wang]{zhang2018unreasonable}
Richard Zhang, Phillip Isola, Alexei~A Efros, Eli Shechtman, and Oliver Wang.
\newblock The unreasonable effectiveness of deep features as a perceptual metric.
\newblock In \emph{Proceedings of the IEEE conference on computer vision and pattern recognition}, pages 586--595, 2018.

\bibitem[Zhang et~al.(2025)Zhang, Jiang, Wei, Chen, Dong, and Yu]{zhang2025generative}
Zhe Zhang, Yili Jiang, Xin Wei, Mingkai Chen, Haiwei Dong, and Shui Yu.
\newblock Generative-ai for xr content transmission in the metaverse: Potential approaches, challenges, and a generation-driven transmission framework.
\newblock \emph{IEEE Network}, 2025.

\end{thebibliography}
}
\clearpage
\setcounter{page}{1}
\maketitlesupplementary

\section{Prompts}
\label{sec:prompt}
Below is the prompt used to instruct VACE to stylize the input video into Tokyo street aesthetics. We deliberately avoid keywords referring to specific landmarks, such as Tokyo Tower, as these cause the model to insert landmark imagery into every generated frame regardless of scene context.

\texttt{In the commercial district, glass curtain walls stand shoulder to shoulder with old European façades; floor-to-ceiling display windows mirror the cherry blossoms blooming on both sides of the street, their pink-and-white shadows rippling at pedestrians’ feet. Following the winding alleyways deeper in, wood-latticed shops accented with vermilion and indigo hang canvas curtains beneath the eaves, and the rain has washed the flagstones to a dark, glossy sheen; a tram thunders across an elevated bridge, its tracks carving an arc through the air, while beneath it tightly packed izakaya and vending machines glow with soft light, waystations for night wanderers. Farther toward the city’s edge, a long greenbelt path runs beside a quietly flowing canal, where rows of Somei-yoshino change color from spring to summer; weeping willows brush the water, and now and then an egret skims past, leaving a ring of fine ripples. In the distance, new high-rises of steel and glass shoot toward the sky, while in the old quarter, wooden machiya still keep their gray tiles and deep eaves, together forming overlapping silhouettes of different eras against the same sky.}

\end{document}